\documentclass{MIPRO}

\usepackage{cite}
\usepackage{amsmath,amssymb,amsfonts}
\usepackage{algorithmic}
\usepackage{graphicx}
\usepackage{textcomp}
\usepackage{xcolor}
\usepackage[T1]{fontenc}
\usepackage{flushend}
\usepackage{multirow}
\usepackage{multicol}
\usepackage{array}
\usepackage{url}

\begin{document}

\title{Bayesian Elicitation with LLMs: Model Size Helps, Extra "Reasoning" Doesn't Always}

\author{
\IEEEauthorblockN{
Luka Hobor,
Mario Brcic,
Mihael Kovac,
Kristijan Poje
}

\IEEEauthorblockA{ 
University of Zagreb Faculty of Electrical Engineering and Computing, Zagreb, Croatia }

luka.hobor@fer.hr, mario.brcic@fer.hr, mihael.kovac@fer.hr, kristijan.poje@fer.hr
}

\maketitle

\begin{abstract}
Large language models (LLMs) have been proposed as alternatives to human experts for estimating unknown quantities with associated uncertainty, a process known as Bayesian elicitation. We test this by asking eleven LLMs to estimate population statistics, such as health prevalence rates, personality trait distributions, and labor market figures, and to express their uncertainty as 95\% credible intervals. We vary each model's reasoning effort (low, medium, high) to test whether more "thinking" improves results. Our findings reveal three key results. First, larger, more capable models produce more accurate estimates, but increasing reasoning effort provides no consistent benefit. Second, all models are severely overconfident: their 95\% intervals contain the true value only 9--44\% of the time, far below the expected 95\%. Third, a statistical recalibration technique called conformal prediction can correct this overconfidence, expanding the intervals to achieve the intended coverage. In a preliminary experiment, giving models web search access degraded predictions for already-accurate models, while modestly improving predictions for weaker ones. Models performed well on commonly discussed topics but struggled with specialized health data. These results indicate that LLM uncertainty estimates require statistical correction before they can be used in decision-making.
\end{abstract}

\renewcommand\IEEEkeywordsname{Keywords}

\section{Introduction}

Bayesian elicitation, i.e., extracting probability distributions or their parameters from domain knowledge, is fundamental across many fields, including medicine, finance, risk assessment, and engineering \cite{o2006uncertain}. Traditionally, this process requires structured interviews with domain experts and is both time-consuming and expensive. With the emergence of large language models (LLMs), researchers have begun exploring whether these models can serve as efficient proxies for probabilistic estimation, given the vast amounts of data they have encountered during training \cite{muller2022transformers}.

Recent LLMs incorporate chain-of-thought (CoT) and extended "thinking" mechanisms that purportedly enable more deliberate reasoning \cite{wei2022chain}. However, multiple studies suggest that CoT may not represent genuine logical reasoning but rather sophisticated pattern matching conditioned on training data distributions \cite{mirzadeh2025gsmsymbolic, zhao2025chain}. Can increased reasoning effort actually improve statistical estimation, or does it merely amplify existing biases and overconfidence?

Prior work has shown that LLMs are systematically overconfident across various tasks \cite{sun2025large, tian2025overconfidence}. LLM-elicited priors are often inaccurate and poorly calibrated \cite{arenda2025openestimate}. Furthermore, extended reasoning has been shown to sometimes impair rather than improve confidence calibration \cite{manikandan2025dontthinktwice}.

In this work, we investigate whether LLMs can serve as Bayesian elicitation assistants and how reasoning effort affects their performance. We construct question sets from four real-world datasets spanning psychology (Big Five personality traits), public health (NHANES, NCD-RisC), and labor markets (Glassdoor). Each question asks models to provide point estimates with 95\% credible intervals for population statistics. We evaluate multiple state-of-the-art models across different reasoning effort levels (low, medium, high) and assess their accuracy, calibration, and interval sharpness.

Our initial results reveal catastrophic under-coverage, reflecting systematic overconfidence in model-generated intervals. To address this, we apply conformal prediction, a distribution-free statistical method that adjusts prediction intervals to guarantee a target coverage rate \cite{angelopoulos2023conformal}, to recalibrate the intervals post-hoc. Based on this investigation, the main contributions of this paper are:
\begin{itemize}
    \item A systematic evaluation of LLMs as Bayesian elicitation assistants across multiple domains, showing that model size drives accuracy while reasoning effort yields no consistent improvement in calibration or accuracy. Against a naive 50\% baseline, models add clear value on familiar topics but struggle on specialized health data.
    \item Empirical evidence that LLM-generated credible intervals exhibit severe under-coverage (overconfidence), with conformal prediction providing an effective post-hoc recalibration mechanism.
    \item A preliminary evaluation of tool-augmented elicitation showing mixed results: web search degrades already-accurate models while providing modest benefits to others.
\end{itemize}

\section{Related Work}

\subsection{Chain-of-Thought Reasoning in LLMs}

Chain-of-thought (CoT) prompting was introduced as a technique to elicit step-by-step reasoning in LLMs, showing substantial improvements on arithmetic and commonsense reasoning benchmarks \cite{wei2022chain}. However, subsequent research has questioned whether CoT represents genuine reasoning or merely sophisticated pattern matching.

Mirzadeh et al. \cite{mirzadeh2025gsmsymbolic} introduced GSM-Symbolic, a benchmark with controlled question variants, finding that LLM performance on mathematical reasoning is unstable. Even semantically equivalent questions with different numerical values or symbols led to accuracy drops of 10-20\%, suggesting a heavy reliance on memorized patterns rather than true reasoning. Zhao et al. \cite{zhao2025chain} examined CoT through a data distribution lens and found that CoT "reasoning" collapses under even moderate distribution shifts, with exact match dropping to 0\% for out-of-distribution transformations. They conclude that CoT is a conditional generation that works only when test cases are similar to training data.

Chlon et al. \cite{chlon2025llms} provide an information-theoretic perspective, showing that transformers are "Bayesian in expectation, not in realization", meaning they approximate Bayesian inference on average across orderings but violate the martingale property for any specific ordering. Interestingly, they derive an optimal formula for chain-of-thought length.

\subsection{LLMs for Bayesian Inference and Uncertainty Quantification}

M\" {u}ller et al. \cite{muller2022transformers} demonstrated that transformers, though not LLMs, can perform Bayesian inference through Prior-Data Fitted Networks (PFNs), learning to approximate posterior predictive distributions in a single forward pass. This foundational work showed that neural networks can learn the mapping from data to Bayesian posteriors without explicit probabilistic computation.

Most closely related to our work, Marzoev et al. \cite{arenda2025openestimate} introduced OpenEstimate, a benchmark for evaluating LLMs on numerical estimation tasks requiring probabilistic priors. Across six frontier LLMs, they found that LLM-elicited priors are often inaccurate and overconfident. Performance improved modestly with different elicitation strategies but was largely unaffected by changes in reasoning effort. Our work complements OpenEstimate by focusing specifically on the interaction between reasoning effort levels and calibration, and by applying conformal prediction to recalibrate model outputs.

\subsection{Overconfidence in Large Language Models}

There are more and more documents pointing to systematic overconfidence in LLMs. Sun et al. \cite{sun2025large} found that all evaluated models show large gaps between accuracy and self-assessed confidence, with overconfidence ranging from 20-60 percentage points. Critically, they observed an inverted Dunning-Kruger pattern: while humans become less biased when uncertain, LLMs become more biased, with confidence declining more slowly than accuracy.

Wen et al. \cite{wen2024mitigating} examined LLM overconfidence through a cognitive psychology lens, finding that larger models exhibit human-like patterns (underconfident on easy tasks, overconfident on hard tasks), whereas smaller models are consistently overconfident. Tian et al. \cite{tian2025overconfidence} identified systematic overconfidence in LLM-as-a-Judge scenarios and proposed the TH-Score metric to better capture calibration in high-stakes evaluation contexts.

Particularly relevant to our investigation, Lacombe et al. \cite{manikandan2025dontthinktwice} found that extended reasoning worsens, rather than improves, confidence calibration. Increasing reasoning budgets caused models to become more overconfident, with accuracy falling sharply while confidence remained high. Again, this suggests that over-reasoning amplifies spurious reasoning patterns rather than improving answer quality.

\subsection{Conformal Prediction for LLM Calibration}

Conformal prediction (CP) provides distribution-free coverage guarantees for prediction sets, making it attractive for LLM uncertainty quantification. Kumar et al. \cite{kumar2023cp} demonstrated CP for multiple-choice QA with LLMs, showing that prediction set size strongly correlates with accuracy and provides better calibration than naive confidence methods. Su et al. \cite{su2024api} introduced LofreeCP, enabling conformal prediction without access to logit values by constructing nonconformity scores from output sampling. That is particularly relevant for API-only models where internal probabilities are unavailable.

\section{Methodology}

\subsection{Datasets and Question Generation}

We construct elicitation questions from four real-world datasets: \textbf{Big Five} personality traits \cite{kajonius2017cross} (conditional probabilities of extreme scores by country and trait), \textbf{NHANES} 2017-2018 health survey \cite{nhanes2018} (prevalence and continuous health metrics by demographic subgroup), \textbf{NCD-RisC} country-level health statistics \cite{ncdrisc2017} (hypertension, diabetes, obesity, BMI by country, sex, and year), and \textbf{Glassdoor} labor market data (job characteristics and salaries by category and location).

For each dataset, we define parameterized question templates whose dimensions vary by domain (e.g., country $\times$ trait $\times$ threshold for Big Five; metric $\times$ country $\times$ sex $\times$ year for NCD-RisC). We enumerate all valid parameter combinations, filter by minimum sample-size thresholds (e.g., $\geq$500 responses per group), and randomly select 100 questions per dataset (400 total).\footnote{\url{https://github.com/LukaHobor/cot_and_bayesian_elicitation}} Here are two question examples: \textit{"Estimate the percentage of people from Germany who are extremely high in Conscientiousness. Provide the percentage and a 95\% confidence interval as three numbers: value, lower, upper."} and \textit{"In 2017--2018, what percentage of Americans answered `yes' to having smoked at least 100 cigarettes in life?"} Crucially, questions ask about \emph{derived statistics} (e.g., conditional percentages, subgroup means) computed from raw data but not appearing verbatim in published sources. This tests whether LLMs can compute these statistics from potentially memorized data rather than recall pre-computed numbers. Ground truth includes appropriate confidence intervals (binomial for proportions, Gaussian for continuous measures).

\subsection{Models and Reasoning Effort Configurations}

We evaluate nine state-of-the-art LLMs spanning different model families and sizes (exact parameter counts and architectures are not publicly disclosed by vendors, so we infer relative capability from each family's product tier):

\begin{itemize}
    \item \textbf{OpenAI:} GPT-5.2, GPT-5-mini, GPT-5-nano
    \item \textbf{Anthropic:} Claude Opus 4.5, Claude Sonnet 4.5, Claude Haiku 4.5
    \item \textbf{Google:} Gemini 3 Pro Preview, Gemini 3 Flash Preview
    \item \textbf{DeepSeek:} DeepSeek-Reasoner
\end{itemize}

Each model is queried at three reasoning effort levels, where supported: low, medium, and high. For models that expose a vendor-defined reasoning effort parameter (e.g., OpenAI's \texttt{reasoning\_effort}, Anthropic's \texttt{extended\_thinking}), we use the API setting directly. For models without such a parameter (e.g., DeepSeek-Reasoner), we control effort via the thinking-token budget: low$=$2{,}000, medium$=$8{,}000, and high$=$16{,}000 tokens. Each question uses a domain-specific prompt template tailored to the dataset (e.g., percentage estimation for Big Five, prevalence or mean estimation for NHANES/NCD-RisC, job-market statistics for Glassdoor). Additionally, two non-reasoning models (\textbf{GPT-4.1} and \textbf{DeepSeek-Chat}) were included as control models that receive no thinking budget. In a separate tool-augmented experiment, we enable web search for six models (Gemini Pro/Flash, Claude Sonnet, GPT-5.2, GPT-5-mini, GPT-5-nano) on 25 questions per dataset at low effort, comparing predictions on matched questions against the tool-free baseline.

\subsection{Answer Extraction and Validation}

Each prompt requests three numbers: a point estimate (value) and a 95\% credible interval (lower, upper bounds). We extract these using a structured extraction pipeline that parses model outputs for numeric triplets. Responses failing to provide valid triplets (non-numeric, incomplete, or malformed) are marked as invalid. Most of these invalid responses consisted of requests for clarification about the problem specification or additional information, which is appropriate abstention when the model's genuine confidence is low.

\subsection{Evaluation Metrics}

\textbf{Negative Log-Likelihood (NLL).} We compute NLL under dataset-appropriate distributions: binomial for proportions, Gaussian for continuous measures. NLL measures how probable the ground truth is under the model's predicted distribution, penalizing both inaccurate point estimates and miscalibrated uncertainty.

\textbf{Coverage.} We compute empirical coverage as the fraction of questions where the ground truth falls within the predicted interval:
\begin{equation}
    \text{Coverage} = \frac{1}{n}\sum_{i=1}^{n}\mathbf{1}[l_i \leq y_i \leq u_i]
\end{equation}
where $l_i$ and $u_i$ are the lower and upper bounds of the predicted credible interval, for well-calibrated 95\% intervals, coverage should approach 0.95.

\textbf{Relative Sharpness.} We measure precision using the coefficient of variation of the predicted interval, $\text{CV}_i = (u_i - l_i) / |\hat{y}_i|$, which provides a scale-invariant measure of interval width. Narrower intervals are preferred when coverage is adequate.

\subsection{Conformal Prediction for Recalibration}

Given the observed undercoverage in the raw model outputs, we apply split conformal prediction to recalibrate the intervals. Because target parameters span vastly different scales, from proportions near 0--1 to continuous measures in the thousands, we use \emph{normalized residuals} rather than additive corrections, ensuring the conformal adjustment scales with each prediction's stated uncertainty.

For each combination of model, reasoning effort level, and dataset, we partition responses into calibration (30\%) and test (70\%) sets. On the calibration set, we compute normalized nonconformity scores $s_i = |y_i - \hat{y}_i| / \sigma_i$, where $\sigma_i = u_i - l_i$ is the predicted interval width. The conformal quantile $\hat{q}$ is the $\lceil(n_{\text{cal}}+1)(1-\alpha)\rceil / n_{\text{cal}}$ empirical quantile of these scores. Test set intervals are then $[\hat{y}_i - \hat{q} \cdot \sigma_i, \hat{y}_i + \hat{q} \cdot \sigma_i]$, providing marginal coverage guarantees under exchangeability. Groups with fewer than 15 calibration points are flagged as statistically insufficient.

\section{Results}

\subsection{Response Validity}

Table~\ref{tab:invalid} shows the proportion of invalid responses and median absolute percentage error (MdAPE) by model and reasoning effort. Invalid responses predominantly consisted of explicit refusals to estimate or requests for clarification.

Most models achieve invalid rates below 25\%, with Claude Haiku stable at $\sim$25\% and Gemini Pro at 21.5\% at low effort (dropping to 0\% at high). DeepSeek-Reasoner exhibits a dramatic shift from 88.8\% invalid at low effort to 0\% at high effort due to \emph{thinking budget exhaustion}: at 2{,}000 tokens, the model exhausts its budget on internal reasoning and produces empty outputs. MdAPE reveals that model capability rather than effort determines accuracy: Gemini Pro ($\sim$16--21\%) and Claude Opus ($\sim$23\%) achieve the lowest errors, while smaller models like Haiku and GPT-5-nano remain above 78\%.

\begin{table}[t]
\centering
\small
\caption{Invalid response rates and median absolute percentage error (MdAPE\%) by model and reasoning effort. Non-reasoning controls are listed separately.}
\begin{tabular}{l|rrr|rrr}
\hline
Model & \multicolumn{3}{c}{Invalid\%} & \multicolumn{3}{c}{MdAPE\%} \\
Effort & L & M & H & L & M & H \\
\hline
haiku-4-5 & 25.0 & 25.0 & 24.8 & 89.9 & 92.8 & 91.0 \\
opus-4-5 & \textbf{1.8} & \textbf{1.5} & 1.8 & 23.6 & \textbf{22.9} & 22.9 \\
sonnet-4-5 & 11.5 & 11.5 & 10.2 & 79.6 & 88.6 & 82.4 \\
deepseek-R & 88.8 & 14.5 & \textbf{0.0} & \textbf{5.9} & 25.7 & 24.2 \\
gem-3-flash & 2.0 & 10.0 & 8.5 & 24.1 & 24.1 & 24.4 \\
gem-3-pro & 21.5 & - & \textbf{0.0} & 15.9 & - & \textbf{20.9} \\
gpt-5-mini & 33.5 & 21.2 & 22.2 & 76.5 & 69.4 & 63.5 \\
gpt-5-nano & 23.0 & 13.8 & 14.2 & 88.8 & 82.7 & 77.8 \\
gpt-5.2 & 16.8 & 12.0 & 13.2 & 70.6 & 39.0 & 37.3 \\
\hline
Effort & \multicolumn{6}{l}{Non-thinking} \\
\hline
gpt-4.1 & \textbf{0.0} & - & - & 38.7 & - & - \\
deepseek-C & \textbf{0.0} & - & - & 45.3 & - & - \\
\hline
\end{tabular}
\label{tab:invalid}
\end{table}

\subsection{Accuracy and Sharpness}

\begin{figure}
  \centering
  \includegraphics[width=0.95\columnwidth]{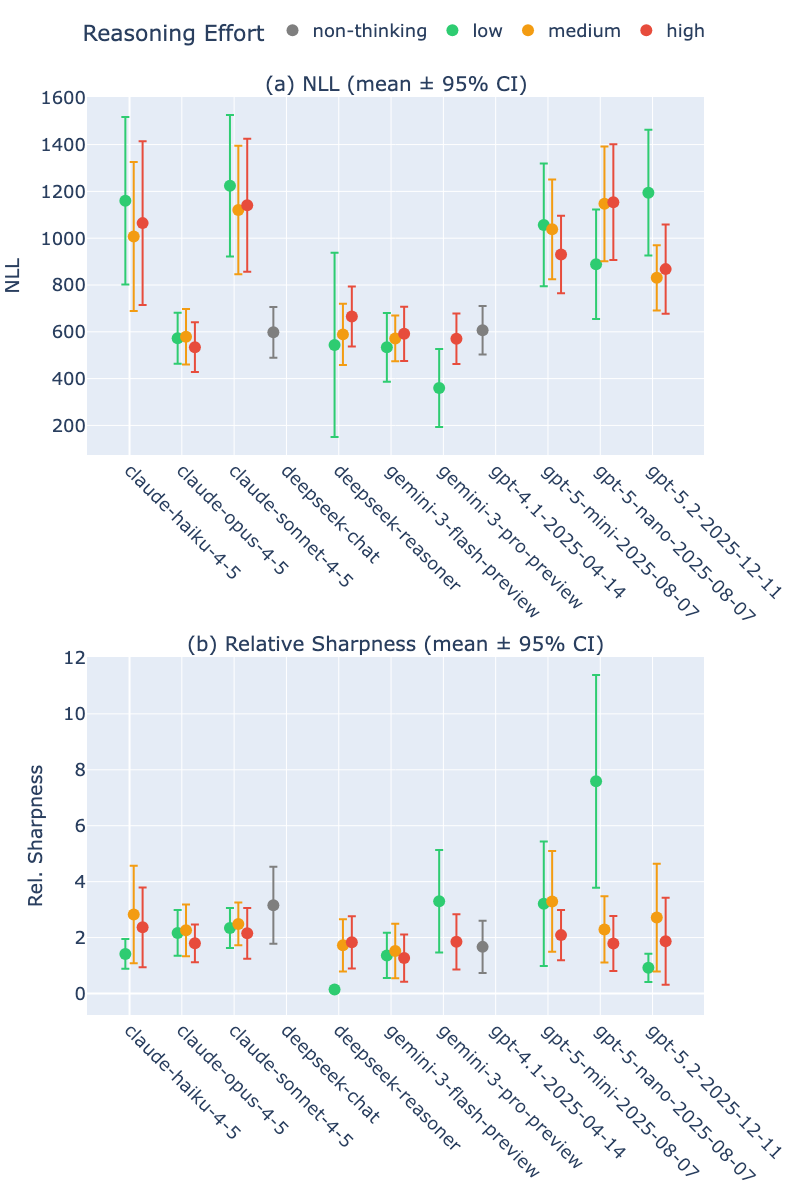}
  \caption{Negative log-likelihood and relative sharpness by model and thinking effort}
  \label{fig:nll_sharp}
\end{figure}

Fig.~\ref{fig:nll_sharp} presents NLL and relative sharpness across models and reasoning effort levels. The dominant factor differentiating performance is model capability rather than reasoning effort. Larger models (e.g., Claude Opus, Gemini Pro) consistently achieve lower NLL than their smaller counterparts. Gemini Pro outperforms Flash overall (median NLL 65 vs.\ 124, $p=0.014$), though at high effort both converge (NLL 130 vs.\ 136, $p=0.78$). Pro's advantage is driven by strong, low-effort performance (median NLL 4.4), with it responding to fewer but easier questions (21.5\% invalid rate).

Per-model Kruskal-Wallis tests reveal that reasoning effort has a statistically significant effect on NLL for four of nine thinking models: GPT-5.2 ($H=9.7$, $p=0.008$), Gemini Flash ($H=7.7$, $p=0.021$), Gemini Pro ($H=31.1$, $p<0.001$), and DeepSeek-Reasoner ($H=17.1$, $p<0.001$). However, the direction of the effect varies: for GPT-5.2, NLL improves from low to medium effort ($r_{rb}=-0.15$, $p_{\text{Bonf}}=0.008$) with diminishing returns at high ($p_{\text{Bonf}}=0.96$). For Gemini and DeepSeek, NLL is paradoxically \emph{lower} at low effort, reflecting selection effects: at low effort, these models respond to fewer questions, and the answered subset tends to be easier. Pooled Spearman trend tests across all thinking models show no significant monotonic relationship between effort and either NLL ($\rho=-0.014$, $p=0.24$) or absolute error ($\rho=-0.013$, $p=0.32$).

\subsection{Coverage and Conformal Calibration}

\begin{figure*}[htbp]
  \centering
  \includegraphics[width=0.85\textwidth]{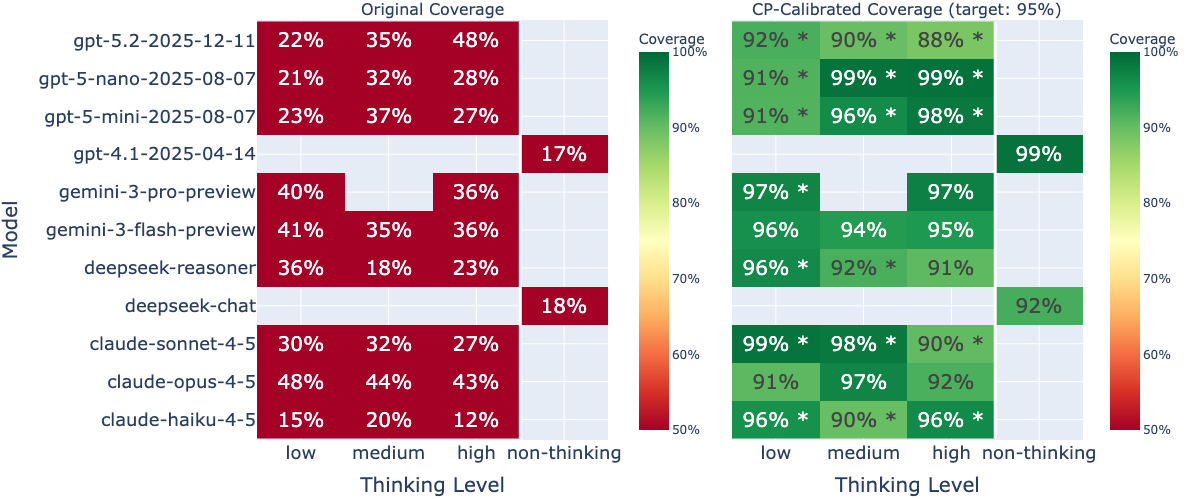}
  \caption{Original and CP-calibrated coverage by model and reasoning effort. Raw coverage (left bars) shows severe under-coverage across all models. Conformal prediction (right bars) recovers near-nominal coverage. Groups marked with * have fewer than 15 calibration points.}
  \label{fig:coverage}
\end{figure*}

Fig.~\ref{fig:coverage} reveals the central finding of this study: all models exhibit severe under-coverage in their raw 95\% credible intervals. Across all model--effort combinations, empirical coverage ranges from 9\% to 44\%, far below the nominal 95\% target. That represents a systematic overconfidence gap.


After applying normalized residual conformal prediction, coverage improves substantially and approaches the 95\% target for most model--effort--dataset combinations. That confirms that the overconfidence is systematic and can be corrected via post-hoc calibration.

\subsection{Aggregated Analysis}

\begin{figure}[htbp]
  \centering
  \includegraphics[width=0.89\columnwidth]{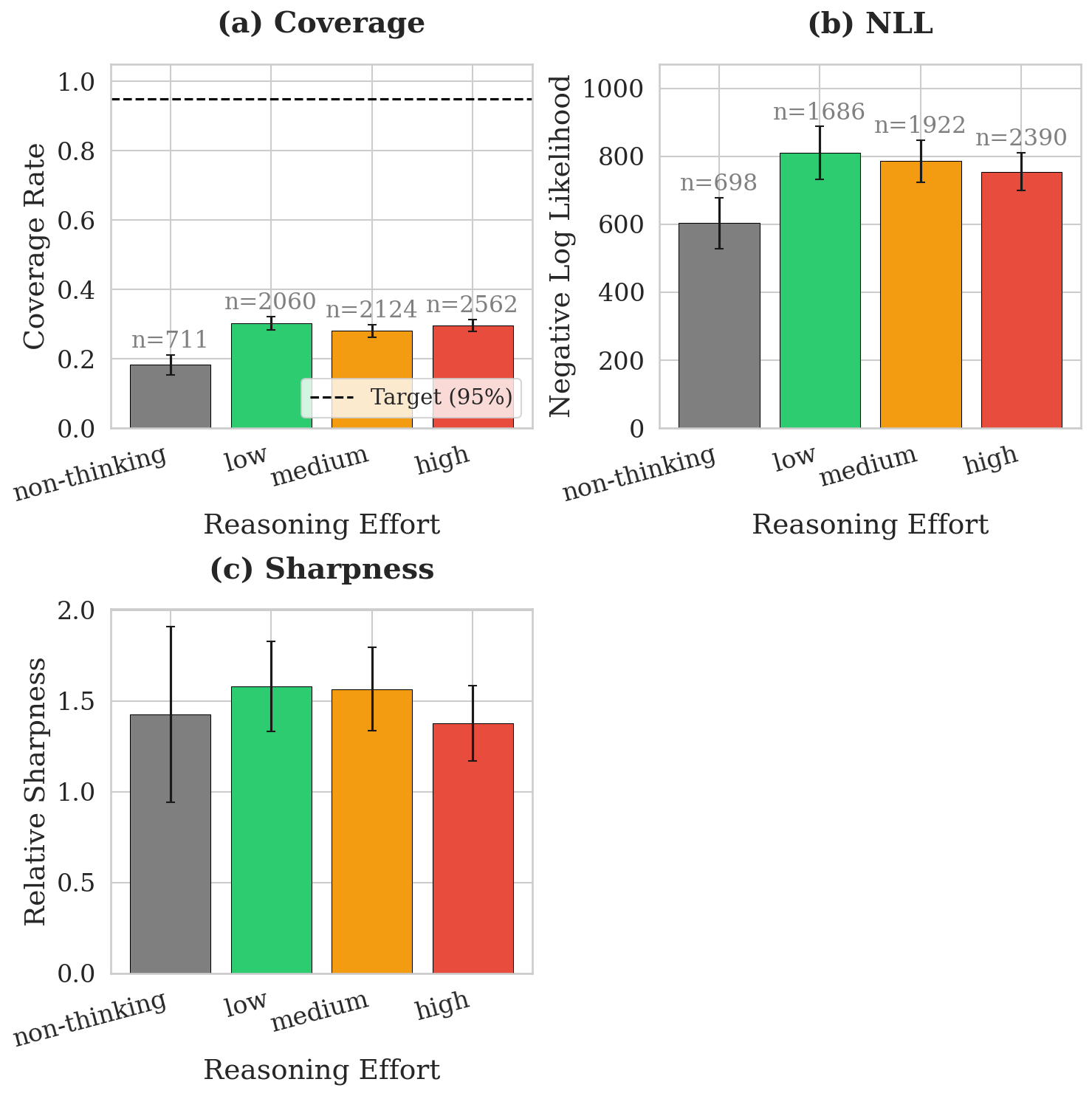}
  \caption{Aggregated NLL, coverage, and relative sharpness by reasoning effort level. NLL and coverage do not vary significantly with effort; only sharpness increases.}
  \label{fig:aggregated}
\end{figure}

Fig.~\ref{fig:aggregated} presents aggregated metrics across reasoning effort levels. When pooling across all thinking models, neither NLL ($H=3.0$, $p=0.22$) nor coverage ($H=2.8$, $p=0.25$) varies significantly with effort level. Only relative sharpness increases with effort ($H=13.5$, $p=0.001$), indicating that models produce wider intervals at higher effort without improving calibration. These results indicate that additional reasoning effort does not translate into better-calibrated uncertainty estimates when averaged across models.

Within-family comparisons confirm that model capability is the primary driver. For Claude (Opus vs.\ Haiku), the larger model wins decisively: lower NLL (134 vs.\ 498, $p<0.001$) and higher coverage (42.6\% vs.\ 9.5\%, $p<0.001$). For GPT-5, the larger 5.2 achieves better coverage (20.0\% vs.\ 14.8\%, $p<0.01$). For Gemini, Pro outperforms Flash overall (NLL 65 vs.\ 124, $p=0.014$), though at high effort both converge ($p=0.78$).

To contextualize accuracy, we compare against a naive 50\% baseline on proportion-type questions. Models decisively outperform this baseline on Big Five (88.9\% win rate) and Glassdoor (65.7\%), with NCD-RisC also favorable (69.6\%). However, on NHANES health data, the advantage is weaker (60.9\%). The failure pattern is systematic: models predict a median of 18.0\% when the true median is 40.3\%, revealing a strong anchoring bias toward low values. For rare conditions (true value $<$20\%), models win 97.5\% of the time, but for common conditions ($>$50\%), this drops to 57.8\%. That suggests models default to predicting low prevalence when uncertain about health statistics, rather than reasoning from population data.

\subsection{Tool-Augmented Elicitation}

We conduct a preliminary experiment, giving four models access to web search during elicitation, and compare against their tool-free baseline on matched questions using Wilcoxon signed-rank tests ($n=25$ questions per dataset, low reasoning effort). Each provider's native search integration was used, with up to 5 searches per query, and the model autonomously deciding when to invoke search.

Overall, web search significantly \emph{degrades} performance ($p=0.009$), with a win rate of only 39.8\%. The effect is asymmetric across models. GPT-5.2, the weakest of the four at low effort, is the only model that benefits: median AE drops from 5.5 to 2.6 with a 65.6\% win rate, though this is not significant ($p=0.21$, $n=32$). In contrast, Claude Sonnet ($p=0.017$) and Gemini Pro ($p=0.006$) are significantly degraded, with web search increasing their median AE by 140\% and 77\%, respectively. Gemini Flash shows no significant change ($p=0.25$). The pattern is also domain-dependent: tools are near-neutral on data-rich topics (Glassdoor: 44\% win rate; NCD-RisC: 47\%) but harmful on personality and health data (Big Five: 38\%; NHANES: 25\%). These results suggest that web search disrupts models that already possess strong internal knowledge while modestly helping those with knowledge gaps.

\section{Discussion}

\subsection{Model Size vs.\ Reasoning Effort}

Our results demonstrate a clear hierarchy: model capability is the primary determinant of elicitation quality, while reasoning effort plays a secondary and inconsistent role. Pooled across all thinking models, neither NLL nor absolute error shows a significant trend with effort level, and four of nine models that show significant per-model effects exhibit mixed directions. That aligns with evidence that chain-of-thought reasoning operates more as conditional pattern matching than genuine deliberation \cite{zhao2025chain, mirzadeh2025gsmsymbolic}. When a model lacks relevant knowledge, extended reasoning cannot compensate- it can only recombine patterns already present in the training distribution \cite{manikandan2025dontthinktwice}.

The domain-dependent results further support this: models excel on commonly discussed topics (Big Five 88.9\%, Glassdoor 65.7\%) but systematically under-predict health statistics, and web search degrades already-accurate models while modestly helping weaker ones.

\subsection{Overconfidence and Conformal Recalibration}

The most striking finding is the universal under-coverage of model-generated credible intervals. No model--effort combination achieved coverage above 44\%, despite all intervals being nominally 95\%. This calibration gap persists across domains and model families, suggesting a fundamental limitation in how LLMs represent uncertainty rather than a model-specific deficiency.

Our findings corroborate prior observations of LLM overconfidence \cite{sun2025large, arenda2025openestimate}. Even the non-reasoning control models exhibit comparable overconfidence, indicating that the problem originates in uncertainty representation rather than the reasoning process itself.

Normalized residual conformal prediction effectively recalibrates intervals to near-nominal coverage. The conformal quantiles $\hat{q}$ typically range from 2 to 5, meaning model intervals must be expanded 2--5$\times$ to achieve proper coverage. However, CP requires a calibration set of labeled examples per model--domain group ($\geq$15 points), limiting its applicability in zero-shot scenarios.

\subsection{Limitations}

Several limitations should be noted. First, models with higher refusal rates may introduce selection bias; models that respond only to easier questions will appear artificially accurate. Second, CP calibration may not generalize to subpopulations beyond our stratification. Third, the tool experiment is preliminary ($n=25$ per dataset, four models). Finally, while we use domain-specific prompt templates, alternative strategies such as few-shot calibration examples or self-consistency via repeated queries could yield different results.

\section{Conclusion}

We evaluated eleven LLMs as Bayesian elicitation assistants across four domains, systematically varying reasoning effort levels and conducting a preliminary tool-augmented experiment. Our experiments reveal three principal findings. First, model capability is the primary driver of elicitation quality, while increased reasoning effort yields no consistent improvement in calibration or accuracy when pooled across models. Second, all models exhibit severe overconfidence, with empirical coverage of nominally 95\% intervals ranging from 9\% to 44\%. Third, normalized conformal prediction effectively recalibrates intervals to near-nominal coverage. A preliminary tool-augmented experiment showed asymmetric effects of web search: while it improved predictions for poorly performing models, it degraded performance for models that were already accurate without tools. That suggests that external retrieval may interfere with internal priors.

Without recalibration, LLM-generated uncertainty intervals are unreliable for decision-making. Practitioners should prefer larger models and apply conformal calibration (Table~\ref{tab:recommendations}). Future work should explore calibration fine-tuning, ensemble methods, and improved tool integration.

\begin{table}[t]
\centering
\small
\caption{Practitioner recommendations for LLM-based Bayesian elicitation.}
\begin{tabular}{p{2.2cm}|p{5.3cm}}
\hline
\textbf{Aspect} & \textbf{Recommendation} \\
\hline
Model choice & Use larger models (Gemini Pro, Claude Opus). \\
\hline
Reasoning effort & Medium effort suffices; high effort provides diminishing or no returns. \\
\hline
Calibration & Always apply conformal prediction. \\
\hline
Sample size & Collect $\geq$15 calibration points per model--domain group for reliable CP. \\
\hline
\end{tabular}
\label{tab:recommendations}
\end{table}

\bibliographystyle{IEEEtran}
\bibliography{references}

\end{document}